\documentclass[10pt,conference,a4paper]{IEEEtran}
\usepackage{ifpdf}
\usepackage{caption}
\ifCLASSINFOpdf
  \usepackage[pdftex]{graphicx}
  \graphicspath{{../pdf/}{../jpeg/}}
  \DeclareGraphicsExtensions{.pdf,.jpeg,.png}
\else
  \usepackage[dvips]{graphicx}
  \graphicspath{{../eps/}}
  \DeclareGraphicsExtensions{.eps}
  \fi
\usepackage{amsmath,amsxtra,amssymb,amsthm,latexsym,amscd,amsfonts}
\usepackage[utf8]{vntex}
\ifCLASSINFOpdf
\usepackage[pdftex]{graphicx}
\DeclareGraphicsExtensions{.pdf,.jpeg,.png,.jpg}
\else
\usepackage[dvips]{graphicx}

\DeclareGraphicsExtensions{.eps}
\usepackage{multirow}
\usepackage{float}
\usepackage{subcaption}
\fi
\usepackage{array}
\usepackage{epstopdf}
\usepackage{anysize}
\marginsize{2.4cm}{2.0cm}{3.2cm}{3.0cm}
\usepackage{multicol}
\usepackage{balance}
\usepackage{graphicx}
\usepackage[font=small]{caption}

\makeatletter
\def\ScaleIfNeeded{\ifdim\Gin@nat@width>\linewidth\linewidth\else\Gin@nat@width\fi}

\begin{document}
\columnsep=0.63cm
\def\mathbi#1{\boldsymbol{#1}}
\def\erfc{\:\mathrm{erfc}}
\def\arg{\:\mathrm{arg}}
\def\E{\:\mathrm{E}}
\def\sinc{\:\mathrm{sinc}}
\def\T{\mathrm{T}}
\def\H{\mathrm{H}}
\newcommand{\bigsize}{\fontsize{16pt}{20pt}\selectfont}

%
\include{Abbr}
\title{Nâng cao hiệu quả ước lượng ảnh chiều sâu cho robot lặn kết hợp xử lý ảnh và học máy}

\author{
\IEEEauthorblockN{
Nguyễn Quang Trường, Nguyễn Cảnh Thanh và Hoàng Văn Xiêm
} 
\IEEEauthorblockA{ Bộ môn Kỹ thuật Robot, Khoa Điện tử - Viễn Thông \\ Trường Đại học Công Nghệ - Đại học Quốc gia Hà Nội\\
		Email: quangtruongnguyen1506@gmail.com, canhthanh@vnu.edu.vn, xiemhoang@vnu.edu.vn}
}
\maketitle

\begin{abstract}
Ngày nay, thông tin độ sâu có vai trò quan trọng trong hệ thống tự trị nhận biết môi trường và ước tính trạng thái của robot. Với sự phát triển nhanh chóng của công nghệ mạng lưới thần kinh sâu, ước tính độ sâu đã được nghiên cứu rộng rãi và chứng minh được tiềm năng ứng dụng trong thực tế. Tuy nhiên, đối với môi trường đặc biệt khó khăn khi cường độ ánh sáng thấp và nhiễu như môi trường dưới nước, việc áp dụng trực tiếp các mô hình học máy có thể không mang lại hiệu quả như mong đợi. Do đó, trong bài báo này chúng tôi trình bày phương pháp cải thiện chất lượng hình ảnh trong môi trường dưới nước nhằm nâng cao hiệu quả ước lượng chiều sâu. Đầu tiên, hình ảnh dưới nước được xử lý thông qua các phương pháp bù màu, cân bằng sáng đồng thời làm tăng độ tương phản, độ nét của các vật thể trong hình. Sau đó, chúng tôi triển khai ước lượng ảnh chiều sâu thông qua mô hình Udepth đối với ảnh đã cải thiện. Cuối cùng, các kết quả được đánh giá và trình bày nhằm kiểm nghiệm hiệu quả cũng như độ chính xác của phương pháp cải thiện chất lượng hình ảnh chiều sâu cho robot lặn.
\end{abstract}

\begin{IEEEkeywords}
Ảnh chiều sâu, cải thiện chất lượng hình ảnh, ước lượng ảnh chiều sâu.
\end{IEEEkeywords}
\IEEEpeerreviewmaketitle  
\section{GIỚI THIỆU}
%
Với sự phát triển của khoa học công nghệ, robot đã dần được ứng dụng vào những nhiệm vụ thám hiểm, đặc biệt là các robot lặn dùng trong nhiệm vụ khám phá đại dương \cite{Shkurti2012}, \cite{Girdhar2014}. Hoạt động trong môi trường đặc biệt như môi trường dưới nước nên robot thám hiểm thường đòi hỏi nguồn thông tin lớn về môi trường, đia hình hay sinh vật. Vì vậy, việc ước lượng hình ảnh chiều sâu trong môi trường dưới nước \cite{Shkurti2012}, \cite{Champion2017} đã được nghiên cứu nhằm cung cấp thêm thông tin cho các nhiệm vụ như tái tạo địa hình 3D, điều hướng điều khiển của robot. Trước đây, các nghiên cứu xây dựng ước tính chiều sâu thường sử dụng các dạng cảm biến như Sonar \cite{Mantani2022} hay máy quét Laser \cite{Palomer2019}. Nhưng với việc chịu ảnh hưởng của nhiễu do hệ thống sinh vật dưới nước hay các phương tiện di chuyển đường biển và các hiện tượng như khúc xạ, phản xạ ánh sáng mà hiệu quả thường khá thấp. Do đó, sử dụng hình ảnh RGB trong việc tính toán chiều sâu môi trường dưới nước ngày càng được quan tâm và có nhiều bước phát triển trong các nghiên cứu \cite{Champion2017}, \cite{Mertan2022}. \cite{Ye2023}. 

Không giống như robot trên mặt đất, robot dưới nước bị hạn chế bởi dẫn đường trực quan do hiện tượng hấp thụ và tán xạ ánh sáng \cite{Schettini2010}. Do đó, việc tăng chất lượng ảnh dưới nước đóng vai trò quan trọng. Các phương pháp nâng cao hình ảnh dưới nước cải thiện chất lượng hình ảnh bằng cách thay đổi giá trị của điểm ảnh. Nhiều phương pháp được mô tả trong các nghiên cứu như phương pháp chỉnh sửa màu sắc của hình ảnh dưới nước thông qua mô hình học không giám sát \cite{Iqbal2010}, phương pháp dựa trên Retinex để hiệu chỉnh màu sắc và tăng cường độ tương phản của hình ảnh dưới nước \cite{Fu2015} hay đề xuất về phương pháp làm mờ hình ảnh dựa trên tích chập đa kênh MSRCR để nâng cao chất lượng hình ảnh dưới nước \cite{Zhang2019}. Các phương pháp cải thiện chất lượng hình ảnh dưới nước trên có những ưu điểm rõ ràng để cải thiện độ tương phản và độ sáng của hình ảnh dưới nước, đồng thời có ưu điểm là nhanh hơn và đơn giản hơn các phương pháp khôi phục hình ảnh dưới nước khác.

Các phương pháp ước lượng chiều sâu sử dụng mono-camera \cite{Godard2017}, \cite{Godard2019} thường là các mô hình hình học dựa trên các cấu trúc chuyển động, stereo thị giác và kết hợp tính năng đa góc nhìn. Các phương pháp này yêu cầu sự tương ứng phù hợp và sức mạnh tính toán đáng kể, tuy nhiên chỉ tạo ra thông tin có chiều sâu thưa thớt. Từ những khó khăn đó, các phương pháp ứng dụng học máy được phát triển \cite{Jiang2022}, \cite{Ye2023}. Các mô hình học máy có thể học cách suy ra các bản đồ độ sâu dày đặc từ các hình ảnh RGB đơn lẻ. Từ đó làm tăng hiệu quả ước lượng thông tin chiều sâu trong hình ảnh. 


\begin{figure*}[!ht]
    \centering
    \includegraphics[width=\textwidth,height=0.4\textwidth]{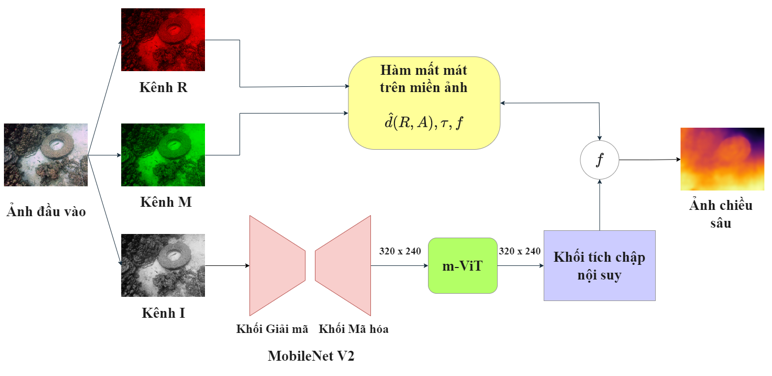}
    \caption{Mô hình ước lượng ảnh chiều sâu Udepth}
    \label{fig:depth_pipeline}
\end{figure*}

Tuy nhiên, những hạn chế về dữ liệu chuẩn trong môi trường dưới nước, độ phức tạp trong quá trình xử lý và ước lượng hình ảnh cũng như những khó khăn về việc ước lượng tỷ lệ ranh giới độ sâu của các đối tượng trong hình ảnh. Để giải quyết những vấn đề trên, bài báo trình bày tập trung vào hai đóng góp chính như sau:

\begin{enumerate}
    \item Đề xuất phương pháp cải thiện chất lượng hình ảnh dựa trên quá trình cân bằng màu sắc và tăng cường độ tương phản, sắc nét cho vật thể trong hình.
    \item Ứng dụng kết hợp mô hình học máy Udepth và quá trình xử lý ảnh trong ước lượng ảnh chiều sâu.
\end{enumerate}


Cấu trúc của bài báo được sắp xếp theo thứ tự như sau: phần II và phần III lần lượt trình bày mô hình ước lượng ảnh chiều sâu và quá trình cải thiện chất lượng ảnh đầu vào dựa trên các thuật toán xử lý ảnh. Phần IV mô tả về tập dữ liệu huấn luyện và những kết quả đã đạt được cũng như so sánh với những mô hình liên quan. Cuối cùng, kết luận và đánh giá được nêu rõ trong phần V.

\section{Mô hình ước lượng ảnh chiều sâu }
\label{Sec:EUDIM}
Trong phần này, chúng tôi trình bày mô hình học máy UDepth \cite{Yu2022} như được minh họa trong Hình \ref{fig:depth_pipeline}. Đây là mô hình sử dụng hình ảnh RGB để ước lượng chiều sâu của các vật thể trong hình từ mô hình tham chiếu đã được huấn luyện. Kiến trúc mạng của mô hình ước lượng ảnh chiều sâu bao gồm ba thành phần chính: Bộ mã hóa (Encoder) và giải mã (Decoder) dựa theo cấu trúc mạng MobileNetV2, Khối biến đổi m-Vision Transformer và Khối tích chập nội suy. Các thành phần này được liên kết tuần tự để thực hiện học có giám sát nhằm ước lượng chiều sâu hình ảnh.

\subsection{Định dạng không gian ảnh đầu vào}
Không gian hình ảnh đầu vào sẽ được chia thành các kênh R - kênh đơn sắc đỏ, kênh M - kênh biểu thị giá trị điểm ảnh lớn nhất giữa kênh đơn sắc xanh lá và xanh dương, kênh I - kênh biểu thị cường độ xám của hình ảnh. Mối tương quan giữa hai kênh R và kênh M là thông tin quan trong trong mô hình học ước lượng ảnh chiều sâu. Kênh I đóng vai trò cung cấp thông tin cường độ xám qua các mạng mã hóa MobileNetV2 và bộ biến đổi mViT để trích suất các thông tin về độ sâu của các điểm ảnh.

\subsection{MobileNetV2 Backbone}
Mô hình ước lượng sử dụng bộ mã hóa và giải mã MobileNetV2. Mạng MobileNetV2 nhanh hơn đáng kể so với các lựa chọn thay thế SOTA khác mà chỉ có ảnh hưởng nhỏ về hiệu suất, điều này giúp cho việc triển khai với các robot hay thiết bị thực tế trở nên khả thi. Bộ mã hóa và giải mã được thế kế theo các lớp có dạng cổ chai ở cuối bộ mã hóa và đầu bộ giải mã. Các lớp mở rộng trung gian sử dụng các phép tích chập theo chiều sâu nhẹ để lọc các đặc điểm như một nguồn phi tuyến tính. Các lớp cuối cùng của bộ giải mã đã được điều chỉnh lớp giải mã tích chập cuối cùng để cuối cùng nó tạo ra 48 bộ lọc có độ phân giải 320 × 480, với đầu vào 3 kênh RMI.

\subsection{Khối biến đổi mViT}
ViT là khối biến đổi Vision được lấy ý tưởng từ kiến trúc Transformer và  các khối MLP (Multilayer Perceptron). ViT mang lại hiệu quả khá tốt khi so sánh với CNN nhưng cũng có hạn chế về kích thước mô hình và độ trễ không phù hợp với các nhiệm vụ yêu cầu độ nhanh và thời gian thực. Do đó, một dạng ViT được phát triển là mViT. Điểm khác biệt của mViT là mỗi patch thông thường sẽ được chia thành 9 m-patch nhỏ hơn , m-patch trung tâm sẽ khai thác thông tin từ các m-patch xung quanh và sẽ đại diện cho patch lớn đó. Các m-patch sẽ đóng vai trò như các patch đầu vào của mViT. Nhờ vậy sẽ làm giảm kích thước mô hình hơn bằng cách sử dụng ưu điểm của khổi Biến đổi và khối Tích chập. Khổi MLP sẽ nhận đầu vào là vectơ trạng thái từ mViT Encoder đưa ra vectơ đặc trưng $f_{b}$.

\subsection{Khối tích chập nội suy}
Cuối cùng, khối tích chập nội suy kết hợp các giá trị tập trung trong phạm vi patch và các vectơ đặc trưng $f_{b}$ để tạo ra hình ảnh đặc trưng $f$. Để tránh sự rời rạc của các giá trị độ sâu, ước lượng của ảnh độ sâu \textbf{D} được tính bằng tổ hợp tuyến tính với giá trị $\overline{f_{b}}$, được biểu diễn trong công thức (\ref{eqn:d_estimate}) như sau: 
\begin{equation}
    \Hat{d}=\Sigma_{k}\overline{f_{b}}_{k}\sigma(R_{k})
    \label{eqn:d_estimate}
\end{equation}

\subsection{Hàm mất mát mục tiêu}
\subsubsection{Hàm bình phương sai số}
Hàm bình phương sai số được sử dụng nhằm thực hiện tối ưu hóa giá trị sai số giữa ảnh ước lượng $\Hat{d_{i}}$ và ảnh tham chiếu $d_{i}$ được mô tả trong công thức (\ref{eqn:least_square_loss}) dưới đây, với N là tổng số lượng giá trị điểm ảnh:
\begin{equation}
    L_{\chi^2}=\frac{1}{N}\sum\limits_{i=1}^{N}(d_{i}-\Hat{d_{i}})^2
    \label{eqn:least_square_loss}
\end{equation}

\begin{figure*}[!ht]
    \centering
    \includegraphics[width=\textwidth,height=0.4\textwidth]{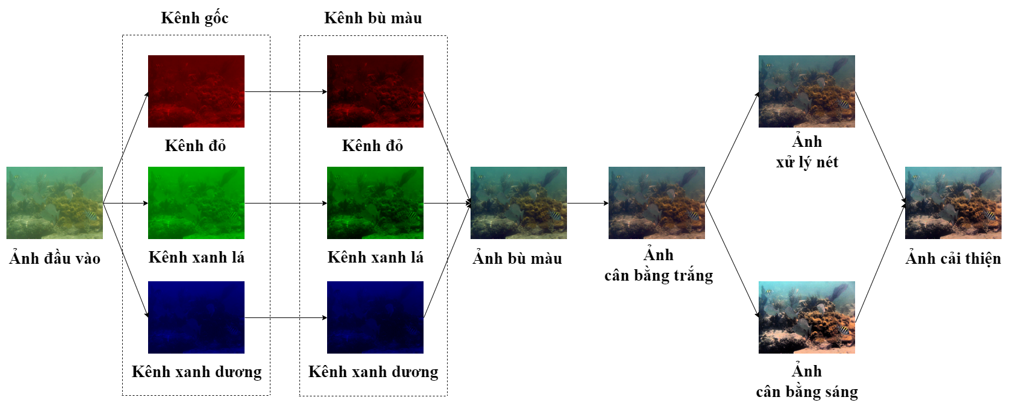}
    \caption{Quá trình cải thiện chất lượng hình ảnh dưới nước}
    \label{fig:enhance_image_pipeline}
\end{figure*}

\subsubsection{Hàm mất mát dữ liệu}
Trong quá trình ước lượng ảnh chiều sâu, thông tin chiều sâu được thu thập chủ yếu trong các vùng hình ảnh vật thể gần với camera và khá thưa thớt trong các vùng không có vật thể hay vật thể ở xa camera. Điều đó có thể gây ra sai sót trong quá trình ước lượng chiều sâu. Để giảm bớt vấn đề mất cân bằng, hàm mất mát dữ liệu đươc trình bày trong công thức (\ref{eqn:data_loss}) thực hiện tính toán sự khác biệt giữa các giá trị độ sâu dự đoán với giá trị ground truth trong miền logarit:
\begin{equation}
    L_{data}(\Hat{d},d)=\alpha\sqrt{
    \frac{1}{N}\sum\limits_{i}t_{i}^2 - \frac{\lambda}{N^2}(\sum\limits_{i}t_{i})^2
    }
    \label{eqn:data_loss}
\end{equation}
Trong đó, $t_{i}=\log{\Hat{d}_{i}}-\log{d_{i}}$. Giá trị cân bằng $\lambda=0.85$ và hệ số $\alpha=10$ được xác định trong nghiên cứu \cite{Bhat2021}.

\subsubsection{Hàm mất mát trên miền ảnh}
Hàm mất mát trên miền ảnh được xây dựng từ mối tương quan giữa kênh màu R-M với giá trị điểm ảnh chiều sâu. Giá trị tương quan $\overline{d}(x|R,M)$ giữa các giá trị điểm ảnh trong hai kênh R và kênh M được mô tả như một hàm tuyến tính trong công thức (\ref{eqn:RM_d}):
\begin{equation}
    \overline{d}(x|R,M)=\tau_{0}+\tau_{1}R(x)+\tau_{2}M(x)
    \label{eqn:RM_d}
\end{equation}
Sau đó thực hiện tối ưu hóa giá trị $\tau$ bằng hàm tối ưu giá trị bình phương nhỏ nhất trong công thức (\ref{eqn:tau}):
\begin{equation}
    \tau=argmin\sum\limits_{i,x}||d_{i}(x)-\overline{d}(x|R_{i},M_{i})||^2_{2}
    \label{eqn:tau}
\end{equation}
Cuối cùng, hàm mất mát trên miền ảnh được trình bày trong công thức (\ref{eqn:domain_loss}):
\begin{equation}
    L_{domain}=E[\prod_{\tau}(\overline{d}_{i}(R,M)-\Hat{d}_{i}]
    \label{eqn:domain_loss}
\end{equation}

Hàm mất mát đưa vào mô hình huấn luyện được tổng hợp từ các hàm mất mát bình phương tối thiểu, hàm mất mát dữ liệu và hàm mất mát trên miền ảnh được mô tả trong công thức (\ref{eqn:loss_function}):
\begin{equation}
    L = \delta_{\chi^2}L_{\chi^2}+\delta_{data}L_{data}+\delta_{domain}L_{domain}
    \label{eqn:loss_function}
\end{equation}
Trong đó , các giá trị $\delta$ được hiệu chỉnh tham số và được tối ưu hóa bằng các giá trị như sau: $\delta_{\chi^2}=0.3$, $\delta_{data}=0.6$ và $\delta_{domain}=0.1$ như trong \cite{Yu2022}.

\section{Cải thiện chất lượng hình ảnh đầu vào trong môi trường dưới nước}
\label{sec:Underwater_Imgae_Enhancement} 
Hình ảnh trong môi trường dưới nước thường có những dặc điểm rất đặc trưng như bị lóa sáng hay tối sáng do hiện tượng khúc xạ, bị biến dạng màu sắc do sự hấp thụ ánh sáng. Những đặc điểm đó đã ảnh hưởng tới hiệu quả ước lượng chiều sâu. Trong phần này, chúng tôi trình bày về phương pháp cải thiện hiệu suất ước lượng ảnh chiều sâu dựa trên phương pháp xử lý ảnh theo cấu trúc trong Hình \ref{fig:enhance_image_pipeline}.

Hình ảnh đầu vào sẽ được chia tách thành ba kênh màu là kênh đỏ, kênh xanh dương và kênh xanh lá, sau đó thực hiện bù màu xanh lá cho các kênh màu đỏ và xanh dương. Tại mỗi điểm ảnh $x$ của kênh đỏ $I_{rc}$ và kênh xanh dương $I_{bc}$ chúng tôi thực hiện bù màu  trong công thức (\ref{eqn:compensate_red}) và (\ref{eqn:compenate_blue}):

\begin{equation}
    I_{rc}(x)=I_{r}(x)+(\bar{I}_{g}-\bar{I}_{r})(1-I_{r}(x))I_{g}(x)
    \label{eqn:compensate_red}
\end{equation}
\begin{equation}
    I_{bc}(x)=I_{b}(x)+(\bar{I}_{g}-\bar{I}_{b})(1-I_{b}(x))I_{g}(x)
    \label{eqn:compenate_blue}
\end{equation}

Trong đó, $I_{r}$, $I_{b}$ và $I_{g}$ biểu thị cho giá trị màu tại điểm $x$ của các kênh đỏ, kênh xanh dương và kênh xanh lá; $\bar{I}_{r}$, $\bar{I}_{b}$ và $\bar{I}_{g}$ là các giá trị trung bình của $I_{r}$, $I_{b}$ và $I_{g}$.\par

Sau khi tiến hành quá trình bù màu cho các kênh đỏ và xanh dương, ảnh sẽ được chỉnh sửa màu và cân bằng trắng bởi thuật toán Gray Wolrd. Thuật toán Gray Wolrd tạo ra ước tính độ chiếu sáng bằng cách tính giá trị trung bình của từng kênh của hình ảnh. Ban đầu, giá trị màu ở các kênh màu đơn sẽ được lấy giá trị trung bình theo như công thức (\ref{eqn:ratio_avg_red}), (\ref{eqn:ratio_avg_blue}) và (\ref{eqn:ratio_avg_green}):\par

\begin{equation}
    r_{R}=\dfrac{max(avg_{r}, avg_{b}, avg_{g})}{avg_{r}}
    \label{eqn:ratio_avg_red}
\end{equation}
\begin{equation}
    r_{B}=\dfrac{max(avg_{r}, avg_{b}, avg_{g})}{avg_{b}}
    \label{eqn:ratio_avg_blue}
\end{equation}
\begin{equation}
    r_{R}=\dfrac{max(avg_{r}, avg_{b}, avg_{g})}{avg_{g}}
    \label{eqn:ratio_avg_green}
\end{equation}

Với các $avg_{r}$, $avg_{b}$ và $avg_{g}$ là các giá trị màu trung bình trong các kênh màu đơn. Sau đó, phương trình (\ref{eqn:r'1c}) và (\ref{eqn:r'2c}) thực hiện quá trình chia thành ba ngưỡng tỉ lệ với $r'^{1}_{c}$ và $r'^{2}_{c}$.\par

\begin{equation}
    r'^{1}_{c} = \alpha_{1}*r_{c}
    \label{eqn:r'1c}
\end{equation}
\begin{equation}
    r'^{2}_{c} = \alpha_{2}*r_{c}
    \label{eqn:r'2c}
\end{equation}

Trong đó, $\alpha_{1}$ và $\alpha_{2}$ là các hằng số và có giá trị tối ưu hóa lần lượt là 0.005 và 0.995. Các giá trị ngưỡng trên được lượng tử hóa bởi hàm lượng tử thấp như được trình bày trong công thức (\ref{eqn:s^{1}_{c}}) và (\ref{eqn:s^{2}_{c}}):\par

\begin{equation}
    s^{1}_{c}=F(I_{c}(x),r'^{1}_{c})
    \label{eqn:s^{1}_{c}}
\end{equation}
\begin{equation}
    s^{2}_{c}=F(I_{c}(x),r'^{2}_{c})
    \label{eqn:s^{2}_{c}}
\end{equation}

Giá trị $I_{c}(x)$ là giá trị màu RGB trong tại điểm ảnh $x$. Để thực hiện quá trình loại bỏ vùng tối mờ hay vùng bóng, trên mỗi kênh màu thực hiện chuẩn hóa giá trị như mô tả trong công thức (\ref{eqn:stand_value}):\par

\begin{equation}
    s_{out}=\left\{
    \begin{array}{lll}
        s^{1}_{c} &    &I_{c}<s^{1}_{c}\\
        I_{c}     &    &s^{1}_{c}\leq I_{c}\leq s^{2}_{c}\\
        s^{2}_{c} &    &I_{c}>s^{2}_{c}
    \end{array}
    \right.
    \label{eqn:stand_value}
\end{equation}

Khi đó, giá trị ảnh đã được cân bằng trắng và chỉnh sửa màu được tính toán theo công thức (\ref{eqn:I^{out}_{c}}):\par
\begin{equation}
    I^{out}_{c} = \frac{s_{out}-s^{1}_{c}}{s^{2}_{c}-s^{1}_{c}}\times255
    \label{eqn:I^{out}_{c}}
\end{equation}

\begin{figure*}[!ht]
    \begin{center}
        \begin{tabular}{ccccc}
             \includegraphics[width=0.18\textwidth]{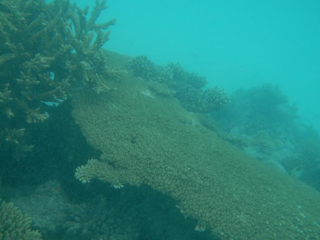}&
             \includegraphics[width=0.18\textwidth]{Depth_Image/Img3_resize.png}&
             \includegraphics[width=0.18\textwidth]{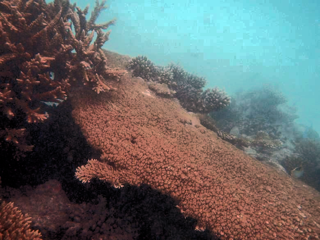}&
             \includegraphics[width=0.18\textwidth]{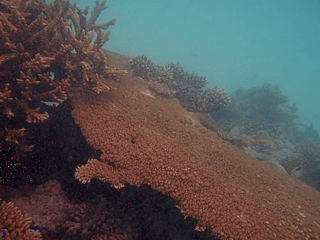}&
             \includegraphics[width=0.18\textwidth]{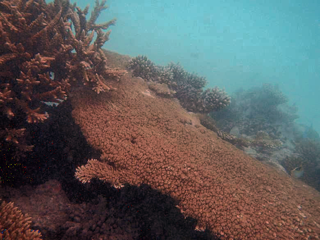}\\
             \includegraphics[width=0.18\textwidth]{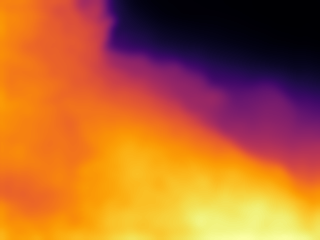}&
             \includegraphics[width=0.18\textwidth]{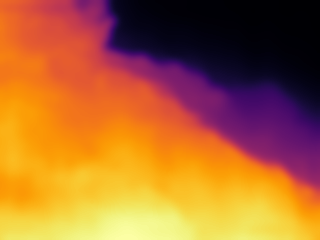}& \hspace{0.01cm}
             \includegraphics[width=0.18\textwidth]{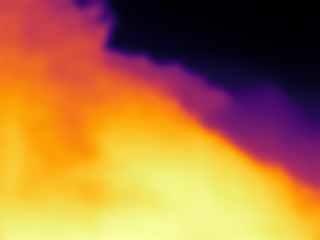}&
             \includegraphics[width=0.18\textwidth]{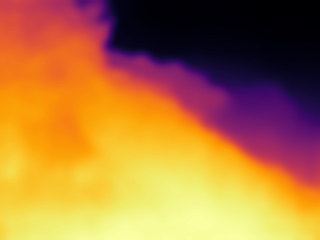}&
             \includegraphics[width=0.18\textwidth]{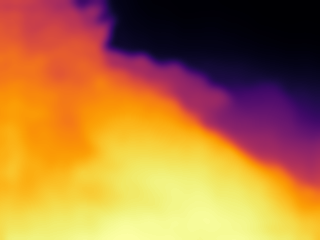}\\
             \includegraphics[width=0.18\textwidth]{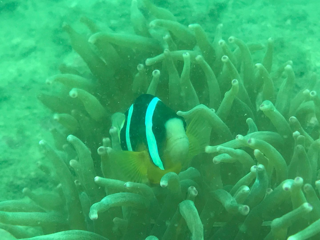}&
             \includegraphics[width=0.18\textwidth]{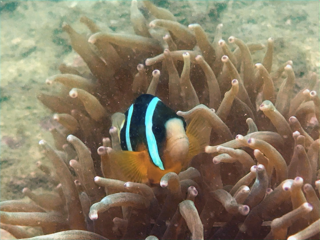}&\
             \includegraphics[width=0.18\textwidth]{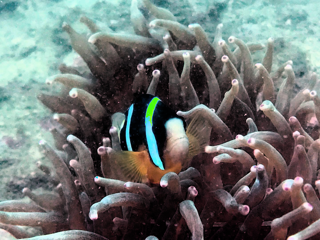}&
             \includegraphics[width=0.18\textwidth]{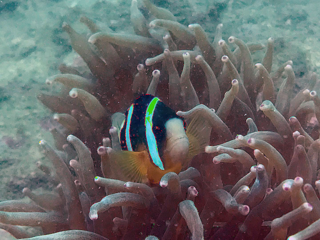}&
             \includegraphics[width=0.18\textwidth]{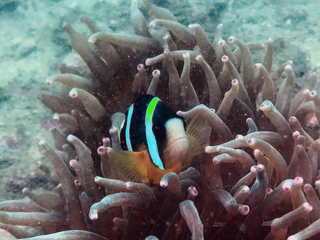}\\
             \includegraphics[width=0.18\textwidth]{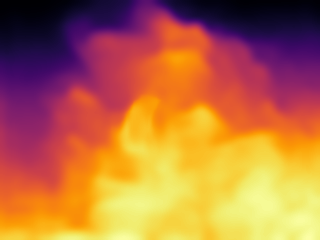}&
             \includegraphics[width=0.18\textwidth]{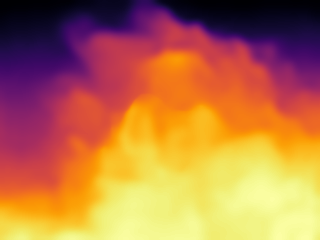}&\
             \includegraphics[width=0.18\textwidth]{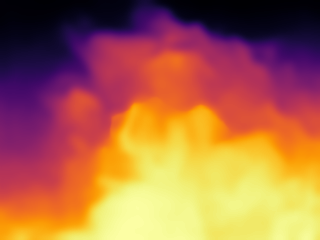}&
             \includegraphics[width=0.18\textwidth]{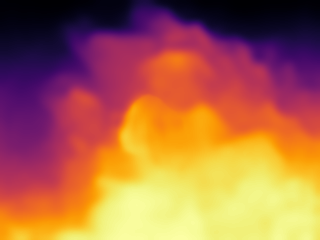}&
             \includegraphics[width=0.18\textwidth]{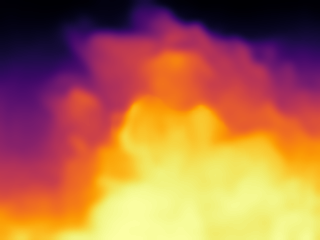}\\
             Ảnh thực & Ảnh WaterNet & Ảnh Histogram & Ảnh Unsharp Marking & Ảnh kết quả\\
             \multicolumn{5}{c}{\includegraphics[width=0.2\textwidth]{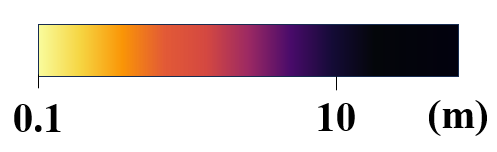}}\\
        \end{tabular}
    \end{center}
    \caption{Ước lượng ảnh chiều sâu trong môi trường dưới nước từ các phương pháp cải thiện chất lượng hình ảnh}
    \label{fig:Depth_Image}
\end{figure*}

Sau đó, hình ảnh được tăng độ tương phản và lọc nhiễu của hình ảnh nhằm tăng độ sắc nét. Độ tương phản được cải thiện bằng phương pháp lượng tử hóa histogram toàn cục (GHE). GHE là một phương pháp ở đó hình ảnh RGB được chuyển đổi sang miền HSV và thực hiện lượng tử hóa thành phần \textit{Value}, giữ nguyên thành phần \textit{Hue} và \textit{Saturation}. Song song với đó quá trình thực hiện tăng độ sắc nét cho hình ảnh của được thực hiện. Quá trình này sử dụng kỹ thuật Unsharp Marking. Công thức (\ref{eqn:Unsharp_Marking}) thể hiện phương pháp Unsharp Marking.\par

\begin{equation}
    I_{UM}(x,y)=2*I(x,y)-GaussianBlur(I(x,y))
    \label{eqn:Unsharp_Marking}
\end{equation}

Trong đó $I(x,y)$ và $I_{UM}(x,y)$ lần lượt là giá trị điểm ảnh tại $(x,y)$ trước khi làm nét và sau khi làm nét. Khi đó, quá trình đã có được hai dạng ảnh là ảnh được tăng độ tương phản và ảnh được tăng độ sắc nét. Cuối cùng, ảnh được cải thiện chất lượng thu được bằng các lấy giá trị trung bình của hai dạng ảnh kể trên theo công thức (\ref{eqn:Avg_Img}) với $I_{fusion}(x,y)$, $I_{GHE}(x,y)$ và $I_{UM}(x,y)$ lần lượt là các giá trị điểm ảnh tại $(x,y)$ của ảnh cải thiện, ảnh tương phản và ảnh lấy nét.

\begin{equation}
    I_{fusion}(x,y)=\frac{I_{GHE}(x,y)+I_{UM}(x,y)}{2}
    \label{eqn:Avg_Img}
\end{equation}

Quá trình cải thiện chất lượng ảnh chiều sâu là quá trình tiền xử lý cho ảnh đầu vào của mô hình ước lượng ảnh chiều sâu. Quá trình này có một vai trò qua trọng việc nâng cao hiệu quả ước lượng và kết quả đánh giá.


\section{KẾT QUẢ}
\label{Sec:Result}
\subsection{Tập dữ liệu huấn luyện}
Trong quá trình huấn luyện, chúng tôi đã sử dụng bộ dữ liệu USOD10K \cite{Hong2023} chứa hình ảnh RGB và hình ảnh chiều sâu tham chiếu cho các cảnh dưới nước khác nhau được chụp ở độ phân giải 640 × 480 điểm ảnh. Bộ dữ liệu chứa 9229 mẫu huấn luyện và 1026 mẫu thử nghiệm bao gồm các ảnh chiều sâu, ảnh RGB và các ảnh mặt nạ của các vật thể.

\subsection{Thông số đánh giá}
Các thông số tiêu chuẩn đánh giá được trình bài trong các công thức dưới đây:
\begin{itemize}
    \item Giá trị trung bình của sai số tương quan tuyệt đối (Abs Rel): 
    $\frac{1}{n}\sum_{p}^{n}\frac{|d_{p}-\Hat{d_{p}}|}{d}$
    \item Sai số bình phương tương quan (Sq Rel): 
    $\frac{1}{n}\sum_{p}^{n}\frac{||d_{p}-\Hat{d_{p}}||^2}{d}$
    \item Trung bình sai số bình phương gốc (RMSE): 
    $\sqrt{\frac{1}{n}\sum_{p}^{d}(d_{p}-\Hat{d_{p}})^2}$
    \item Sai số $log_{10}$: 
    $\frac{1}{n}\sum_{p}^{d}|log_{10}(d_{p})-log_{10}(\Hat{d_{p}})|$
\end{itemize}
Trong đó, $d_{p}$ là giá trị điểm ảnh ground truth chiều sâu $d$, $\Hat{d_{p}}$ là giá trị điểm ảnh của ảnh ước lượng chiều sâu $\Hat{d}$, n là số lượng điểm ảnh trong ảnh tham chiếu cũng như ảnh ước lượng.

\subsection{Kết quả thực nghiệm}
Bộ dữ liệu được cải thiện chất lượng qua quá trình tiền xử lý hình ảnh và huấn luyện với mô hình Udepth với khoảng hơn 16 triệu tham số. Quá trình huấn luyện được thực hiện trên máy trạm có cấu hình gồm CPU Intel Core i3-10105F, RAM Lexar 8G Buss 3200/DDR4, card đồ họa Asus Tuf GTX 1060 Super và mạch chính MSI H510M. Thời gian huấn luyện trung bình của mỗi chu trình là 26 giờ. Các kết quả được so sánh với phương pháp cải thiện chất lượng hình ảnh WaterNet, phương pháp lượng tử Histogram và phương pháp làm nét hình ảnh Unsharp Marking.

\begin{table}[ht]
    \centering
    \caption{Kết quả ước lượng ảnh chiều sâu}
    \begin{tabular}{|c|c|c|c|c|}
    \hline
                       & Abs Rel & Sq Rel & RMSE  & $log_{10}$ \\ \hline
    Ảnh thực           & 1.379   & 0.382  & 0.376 & 0.278  \\ \hline
    Ảnh WaterNet       & 0.745   & 0.176  & 0.174 & 0.219  \\ \hline
    Ảnh Histogram      & 0.621   & 0.165  & 0.159 & 0.234  \\ \hline
    Ảnh Unsharp Marking& 0.767   & 0.204  & 0.232 & 0.256 \\ \hline 
    Ảnh kết quả        & \textbf{0.598} & \textbf{0.132} & \textbf{0.126} & \textbf{0.186} \\ \hline
    \end{tabular}
    \label{Tab:resultsystem}
\end{table}

Từ Bảng \ref{Tab:resultsystem} cho thấy kết quả của mô hình kết hợp tốt hơn với các giá trị tham số nhỏ hơn từ 8\% - 10\%. Giá trị sai số tuyệt đối Abs Rel và sai số tương quan Sq Rel cho thấy ảnh kết quả của phương pháp đề xuất tốt hơn giữa các chi tiết vật thể trong ảnh. Giá trị trung bình sai số RMSE thể hiện ảnh kết quả rõ ràng hơn và có giá trị chiều sâu cao hơn so với các ảnh còn lại. 

Trong hình \ref{fig:Depth_Image}, hình ảnh thực bị ảnh hưởng của biến dạng màu sắc nên ảnh chiều sâu xuất hiện các cạnh mờ và không rõ các vật thể nhỏ. Hình ảnh của phương pháp đề xuất có ưu điểm là độ sắc nét hơn so với ảnh WaterNet và Unsharp Marking và có độ tương phản tốt hơn ảnh Histogram, thể hiện ranh giới độ sâu rõ ràng của các chi tiết vật thể. Từ đó cho thấy, quá trình cải thiện chất lượng hình ảnh đã giúp nâng cao hiệu quả ước lượng chiều sâu của mô hình.

Bên cạnh đó, do cần cải thiện chất lượng hình ảnh dưới nước nên khi kết hợp với mô hình ước lượng ảnh chiều sâu sẽ làm tăng thời gian tính toán ước lượng và tăng độ phức tạp của cả quá trình. Cùng với đó, những hạn chế về dữ liệu hình ảnh trong các môi trường dưới nước cũng như độ chính xác của dữ liệu hình ảnh chiều sâu tham chiếu cho quá trình huấn luyện đã làm giảm đi hiệu quả ước lượng chiều sâu và khả năng ứng trong nhiều môi trường nước khác nhau. Đây cũng là những tiền đề để mở ra những hướng cải thiện mới cho mô hình ước lượng ảnh chiều sâu.

\section{KẾT LUẬN}
\label{Sec:KetLuan}
Trong bài báo, chúng tôi đề xuất phương pháp cải thiện chất lượng hình ảnh dưới nước dựa trên việc bù màu, cân bằng sáng và tăng độ tương phản, đô sắc nét cho hình. Phương pháp xử lý hình ảnh dưới nước trên được kết hợp với mô hình học máy ước lượng ảnh chiều Udepht đã cho thấy những kết quả khả quan trong việc ước lượng ảnh chiều sâu. Trong tương lai, chúng tôi sự định ứng dụng mô hình học sâu trong quá trình cải thiện chất lượng hình ảnh dưới nước và nâng cao độ hiệu quả của mô hình ước lượng ảnh chiều sâu để có thể sử dụng cho robot lặn trong thời gian thực.  

\bibliographystyle{IEEEtran}
\balance
\bibliography{reference}
\end{document}